\documentclass{article}
\usepackage[final]{neurips_2024}
\usepackage[utf8]{inputenc}
\usepackage[T1]{fontenc}
\usepackage{hyperref}
\usepackage{url}
\usepackage{booktabs}
\usepackage{amsfonts}
\usepackage{nicefrac}
\usepackage{microtype}
\usepackage{xcolor}
\usepackage[frozencache,cachedir=.]{minted}
\usepackage{graphicx}
\usepackage{plantuml}
\usepackage{tcolorbox}
\usepackage{listings}
\usepackage{amsmath}
\usepackage{tcolorbox}
\usepackage{amsmath}
\usepackage[capitalise]{cleveref}
\usepackage{subcaption}
\usepackage{tikz}
\usepackage{svg}

\usetikzlibrary{shapes.geometric, arrows.meta, positioning}
\tcbset{
    userquery/.style={colback=blue!5!white, colframe=blue!75!black, fonttitle=\bfseries},
    agentcall/.style={colback=green!5!white, colframe=green!75!black, fonttitle=\bfseries},
    summary/.style={colback=red!5!white, colframe=red!75!black, fonttitle=\bfseries},
}

\definecolor{bg}{rgb}{0.95,0.95,0.95}

\title{LLM Agent for Fire Dynamics Simulations}

\author{%
  Leidong Xu\thanks{Work performed during internship at FM.} \\
  University of Connecticut\\
  \texttt{leidong.xu@uconn.edu} \\
  \And
  Danyal Mohaddes\thanks{Corresponding author.} \\
  FM, Research Division\\
  \texttt{danyal.mohaddes@fmglobal.com} \\
  \And
  Yi Wang \\
  FM, Research Division\\
  \texttt{yi.wang@fmglobal.com} \\
}

\begin{document}

\maketitle

\begin{abstract}
Significant advances have been achieved in leveraging foundation models, such as large language models (LLMs), to accelerate complex scientific workflows. In this work we introduce FoamPilot, a proof-of-concept LLM agent designed to enhance the usability of FireFOAM, a specialized solver for fire dynamics and fire suppression simulations built using OpenFOAM, a popular open-source toolbox for computational fluid dynamics (CFD). FoamPilot provides three core functionalities: code insight, case configuration, and simulation execution. Code Insight is an alternative to traditional keyword searching leveraging retrieval-augmented generation (RAG) and aims to enable efficient navigation and summarization of the FireFOAM source code for developers and experienced users. For case configuration, the agent interprets user requests in natural language and aims to modify existing simulation setups accordingly to support intermediate users. FoamPilot's job execution functionality seeks to manage the submission and execution of simulations in high-performance computing (HPC) environments and provide preliminary analysis of simulation results to support less experienced users. Promising results were achieved for each functionality, particularly for simple tasks, and opportunities were identified for significant further improvement for more complex tasks. The integration of these functionalities into a single LLM agent is a step aimed at accelerating the simulation workflow for engineers and scientists employing FireFOAM for complex simulations critical for improving fire safety.
\end{abstract}

\section{Introduction}

\begin{figure}
    \centering
    \begin{subfigure}[b]{0.43\textwidth}
        \includegraphics[width=\textwidth]{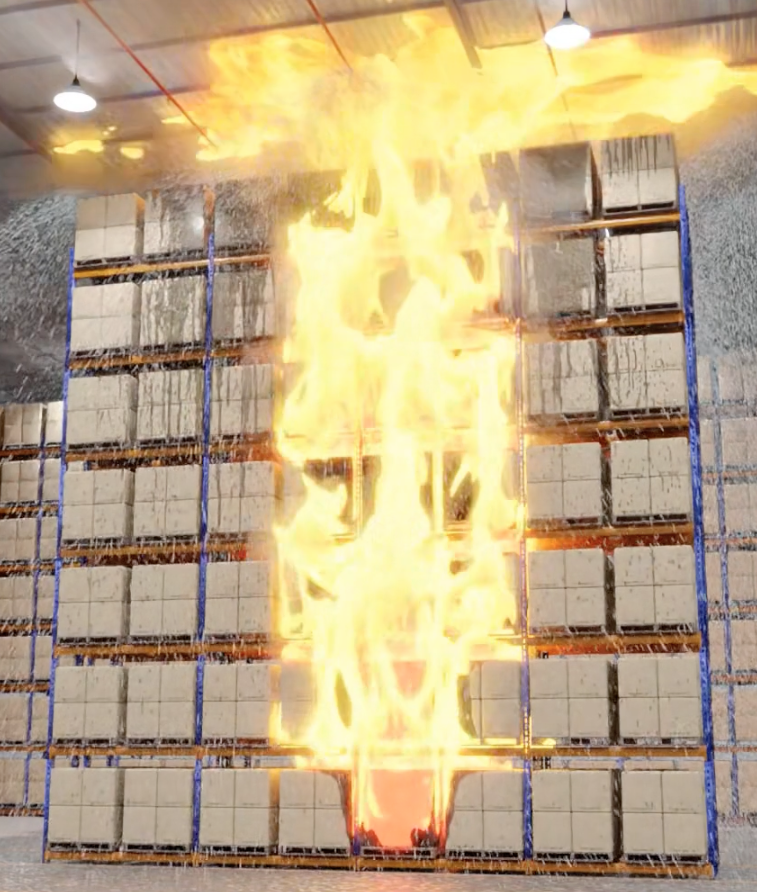}
        \phantomcaption
        \label{subfig:case_render}
    \end{subfigure}%
    \begin{subfigure}[b]{0.55\textwidth}
        \begin{minted}[bgcolor=bg, frame=lines, fontsize=\small, escapeinside=||]{python}
Case/
  mesh.sh # Mesh generation script
  system/ # Simulation and geometry setup
    controlDict # Runtime and I/O control
    extrudeToRegionMeshDict.fuel # Solid mesh
    snappyHexMeshDict # Geometry
    topoSetDict # Domain boundaries
    ...
  constant/ # Sub-model parameters
    radiationProperties
    turbulenceProperties
    ...
    solidRegion/ # Parameters for subdomains
      radiationProperties
      ...
  0/  # Initial and boundary conditions
    T
    U
    ...
        \end{minted}
        \phantomcaption
        \label{subfig:case_structure}
    \end{subfigure}%
    \caption{Left: render of a FireFOAM simulation of large-scale fire suppression (\cite{REN2017695}). Right: typical FireFOAM case structure.}
    \label{fig:case_and_render}
\end{figure}

Computational fluid dynamics (CFD) is an indispensable tool in the study and simulation of fire dynamics and combustion phenomena (\cite{REN2017695, NMIRA2022122746}).
Among the various CFD solvers, FireFOAM\footnote{Available: \url{https://github.com/fireFoam-dev/fireFoam-v1912}} (\cite{WANG20112473}),
a solver developed specifically for fire dynamics simulations,
stands out for its ability to model complex fire and fire suppression physics, including buoyancy-driven turbulence, gas-phase chemical reactions, solid-phase heat transfer and pyrolysis, liquid film flow and spray transport. 
FireFOAM is built using OpenFOAM\footnote{Available: \url{https://www.openfoam.com/}}, a popular C++-based open-source toolbox for CFD.
A rendered image of a typical FireFOAM simulation of a large-scale fire suppression scenario in a warehouse is shown in \cref{subfig:case_render}.
Despite its advanced capabilities, the process of setting up, running and post-processing FireFOAM simulations remains a challenging and time-consuming task, especially on high-performance computing (HPC) architectures.
Users must navigate a multitude of keywords, configurations, and code parameters to achieve accurate and reliable results, creating a steep learning curve for entry-level users and remaining time-consuming even for experienced users. 
A high-level overview of a typical FireFOAM case structure is shown in \cref{subfig:case_structure}.
This complexity necessitates a level of expertise not only in fire science, but also in CFD and HPC, creating a significant barrier for many researchers and engineers.

The development of foundation models, especially large language models (LLMs), has revolutionized various scientific and technical fields by offering advanced capabilities in understanding and generating human-like text and code. 
LLMs, such as OpenAI's GPT series, Meta's LLaMA and Anthropic's Claude, and specialized models like OpenAI's earlier Codex project (\cite{openai2021}) and Meta's Code LLaMA (\cite{meta2023}), have demonstrated substantial capabilities in natural language processing, code generation, and automated problem-solving by interpreting natural language queries to generate corresponding code and thereby aiding software development (\cite{brown2020language, roziere2023code, feng2020codebert, achiam2023gpt}).

While LLMs can generate code from natural language descriptions, it is typically a static output, namely a string of text. The LLM lacks the ability to interact with its computational environment, adapt to changing requirements, or handle unexpected errors autonomously (\cite{jiang2024survey, Wang_2024}). 
To overcome these limitations, an LLM-based agent system is needed—one that not only generates code, but also iteratively tests, debugs, optimizes, and integrates it within a broader software ecosystem (\cite{liu2024large, hong2023metagpt, qian2024chatdev, bouzenia2024repairagent}). 
Such systems combine the strengths of LLMs with additional tools for automated code execution, validation, and refinement, enabling a more comprehensive approach to software development.

Previous works have leveraged LLM agents to automate scientific work (\cite{xia2024llm}). 
Indeed, some works on supporting scientific and engineering simulations, such as~\cite{chen2024metaopenfoamllmbasedmultiagentframework}, have shown success in modifying parameters of existing simulation configurations.
Parameter modification, and more generally simulation case setup, is an important aspect of the scientific simulation workflow.
Considered more broadly, the scientific simulation workflow typically involves problem identification and parametrization, simulation case setup, simulation execution, post-processing of output data and extraction of the desired physical insights.
In this work, we introduce FoamPilot, an LLM agent designed to address three core functionalities within the scientific simulation workflow for both new and experienced human users of FireFOAM: Code Insight, Case Configuration, and Job Execution.

\begin{itemize}
  \item Code Insight: Allow users to quickly locate relevant sections of source code, either to understand its functionality or make necessary modifications.
  \item Case Configuration: Interpret user requests expressed in natural language to accurately create and modify simulation setups.
  \item Job Execution: Manage the submission and execution of simulations in HPC environments and provide preliminary analysis of results.
\end{itemize}

While the present work will focus on FoamPilot and its development for supporting FireFOAM simulations, it should be noted that the challenges identified and the solution approaches developed are broadly applicable to other scientific simulation tools, particularly large, open-source, command-line-based simulation codes.

\section{The FoamPilot agent\label{sec:agent_overview}}

\begin{figure}[h]
   \centering
   \includegraphics[width=\textwidth]{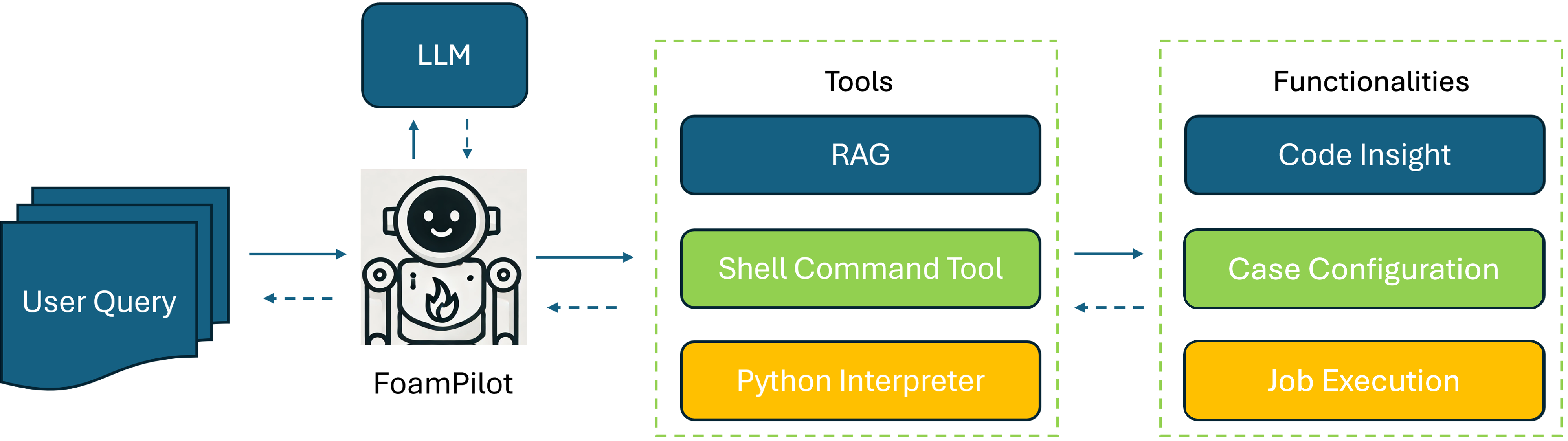}
   \caption{The structure of the FoamPilot agent. Solid and dashed lines indicate queries and feedback, respectively.}
   \label{fig:framework}
\end{figure}

An overview of the FoamPilot agent is illustrated in \cref{fig:framework}. 
The agent is implemented using the LangChain/LangGraph 0.2 framework (\cite{Chase_LangChain_2022}) and is designed to be agnostic to the choice of large language model (LLM). 
This design allows for flexibility in using local or cloud-hosted, open-source or closed-source LLMs, enabling the agent to leverage the ongoing performance improvements in these models, such as those trained for complex reasoning tasks (e.g., o1 from \cite{openai2024}, still in limited preview at the time of writing).

The agent structure follows a graph consisting of three nodes, as shown in \cref{fig:framework}: user, LLM, and tools, with edges connecting them to facilitate message transfer.
The tool node provides access to utilities, which in the present work consist of a Shell Command Tool, a Python Interpreter Tool, and a Retrieval-Augmented Generation (RAG) tool.
These tools were chosen due to their direct utility for the agent's desired functionalities.
The Shell Command Tool executes Linux commands, the Python Interpreter Tool runs Python scripts generated by the LLM, and the RAG Tool identifies the embeddings closest to the query to retrieve information from the database, which, in this work, is a vector store of the FireFOAM source code. Further tools could have been added in the design of the agent, but this was decided against to avoid giving the agent redundant or unnecessary capabilities which could result in mistaken tool choices and accordingly diminished performance.

The LLM processes user queries in natural language augmented with information regarding available tools.
The agent then dynamically and iteratively produces a structured output flow to engage with the available tools to achieve the user's desired functionalities. In our implementation, a static conditional edge, containing "if" statements to direct the flow to different nodes based on the LLM's structured output, is used to invoke the appropriate tools from the tool node and to determine when to terminate the process. The tools cannot interact directly, but can be called by the LLM in the same query. The LLM must orchestrate their use through potentially multiple loops of query and feedback.
The process continues until the task is successfully completed, the LLM gives up, or the context window limit is approached, at which point a summary is provided to the user for progress tracking and feedback purposes.

\subsection{Code Insight\label{subsec:code_insight}}

A user of a simulation code, particularly a developer-user, will often need to refer to the source code to understand the details of sub-model implementations, as well as how to invoke those sub-models in the simulation configuration files.
Traditional methods for navigating and understanding source code often rely on basic text search tools like `grep' and `find'.
These methods can be inefficient and cumbersome because variable names and in-line comments depend on the code author's personal style. 
In large open-source projects like OpenFOAM and FireFOAM, this can turn code navigation into a time-consuming task.
Furthermore, translating classes and functions back to their original mathematical representations, which is often necessary when seeking to understand complex code that is sparsely commented and documented, can also be time-consuming.

RAG is a natural language processing (NLP) approach that enhances text generation by integrating external knowledge retrieval to improve factual accuracy, coherence, and context relevance.
Using RAG with an embedding space search based on cosine similarity offers a significant advantage over traditional user-driven keyword-based searches like `grep' and `find' for navigating FireFOAM's C++ source code. 
Unlike manual keyword matching, the RAG method captures semantic relationships and contextual relevance, enabling it to identify conceptually related code fragments even if they do not share exact terms. 
This typically leads to more meaningful search results, allowing users to find relevant code more efficiently.

In preprocessing the FireFOAM source code for use with the RAG Tool, we have developed a method to address the challenges posed by the separation of C++ header files (.H) and source files (.C). 
Header files primarily contain variable declarations and function prototypes, lacking details on their actual implementation, which are provided in the source files.
If these files were indexed separately, a retrieval model might mostly return header files, providing limited value for the downstream LLM. 
To improve the usefulness of the retrieved content, we combine the header and source files into a single document before passing them to an embedding model.
This approach ensures that both the declarations and the corresponding implementations are available in the same context, making the retrieved information more valuable and comprehensive.
It is important to note that popular embedding models typically have smaller maximum input sizes than the context windows of the most performant LLMs. 
The standard practice is to split or `chunk' large documents into smaller pieces for embedding.
However, most of FireFOAM's header and source files do not exceed the maximum context window of the embedding model in this work (8192 tokens). 
Additionally, placing the header file first in the combined context ensures that the embedded vectors always include critical information from key terms. Thus, even if some of the source code exceeds the embedding model's maximum input size, it can simply be truncated and the similarity search in embedding space still performs well. The correct file is returned to the LLM, the entirety of which comfortably fits within the LLM's context window. This allows for better continuation during complex task execution, since even after multiple tool calls, the source code files will still be present within the context window. This also demonstrates the importance of large context windows for keeping track of the details of long chains of LLM-tool interactions.

\begin{figure}[h]
   \centering
   \includegraphics[width=\textwidth]{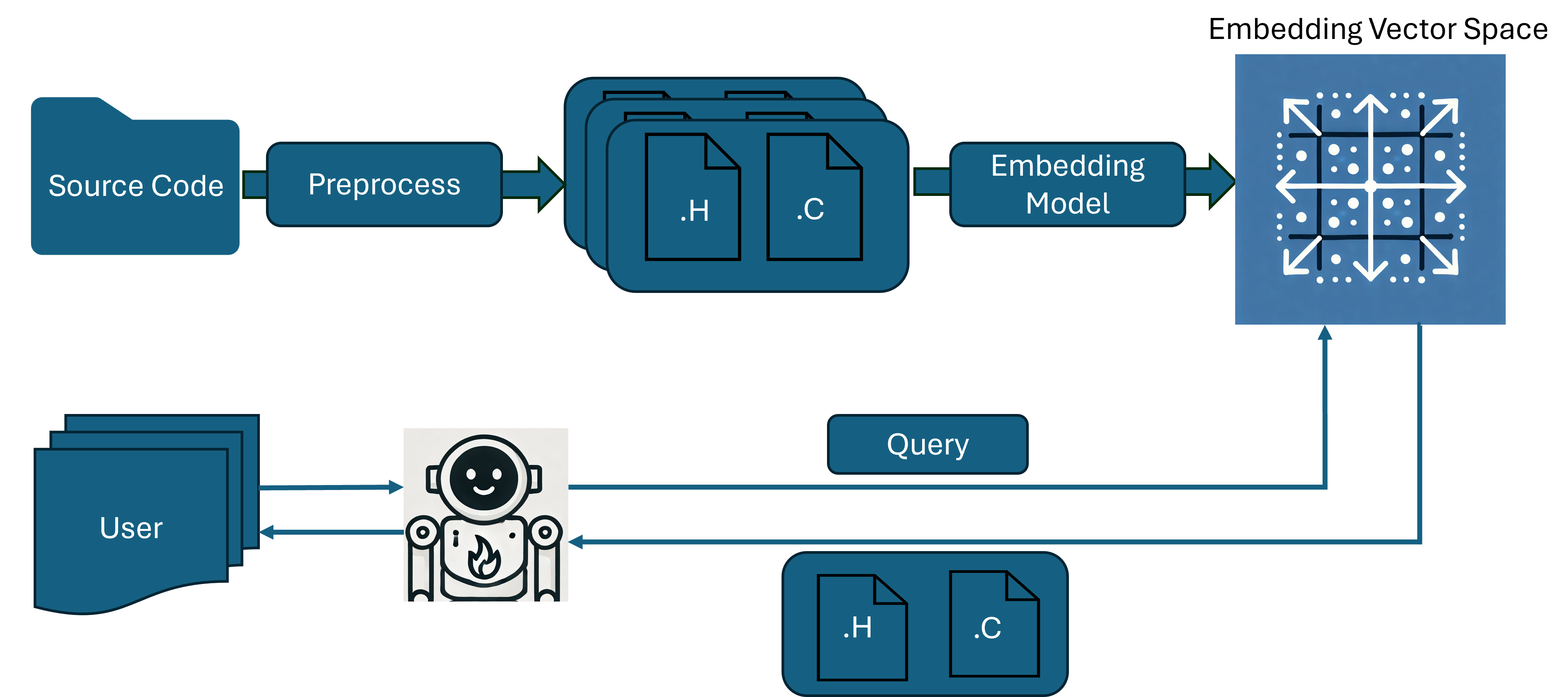}
   \caption{Illustration of the process by which the source code is embedded in a vector database and retrieved by the RAG Tool.}
   \label{fig:rag}
\end{figure}

Additional preprocessing steps included prepending the relative file path to each combined file, providing contextual cues that can be leveraged by the LLM upon retrieval to better understand the organization of the codebase. 
To optimize token usage and remove unnecessary content, we stripped out boilerplate headers and license information, as these elements do not contribute meaningful information for code understanding or search tasks. Comments in the headers and throughout the code were retained, as these provide useful contextual information. After these steps, the preprocessed data were embedded, with similarity search and retrieval accomplished using a vector store. The preprocessing and retrieval steps are illustrated in \cref{fig:rag}.
Compared to the manual keyword search approach, this method is contextually aware, improving the effectiveness of code search within the FireFOAM codebase.

\subsection{Case Configuration}

FoamPilot interprets user requests for case configurations expressed in natural language. Currently, it is capable of modifying an existing simulation case provided by the user.
Initially, we sought only to point the agent to the location of an existing case, allowing it to autonomously find and choose the necessary configuration files to modify to meet the user's request. We found that this was an inefficient approach, with the agent issuing calls to the Shell Tool to individually check every file within the case folder. The case folder for a FireFOAM simulation contains many individual files, as shown in \cref{subfig:case_structure}. 

To address this, we applied an approach similar in spirit to the RAG Tool described in \cref{subsec:code_insight}: 
we stripped out all boilerplate headers and license information, 
prepended the relative file path to each configuration file,
then compressed the entire case folder into a single long string. 
We then provided the case configuration string to the agent within the prompt.
For FireFOAM simulations, the total token usage for this case configuration string was approximately 20k tokens using the GPT-4o tokenizer, well below that model's context window limit of 128k. 
The case configuration string provides the agent with a comprehensive overview of the case, allowing it to efficiently identify the relevant parameters and locations within the configuration files. 
The case configuration prompt is shown in \cref{fig:case_configuration_prompt}, where \texttt{user\_request} is the user's description of their desired modification to the case they have provided.

We contend that the case configuration functionality makes FireFOAM more accessible to non-expert users, by reducing the complexity and associated time required to correctly configure simulations.
Our approach assumes the user possesses a case that they wish to modify according to some request, but does not require the user to provide additional example cases.
This is a typical situation for a junior scientist, who may inherit a simulation case from a senior scientist and then be required to modify it according to some request, but may not have access to a database of example cases relevant to the specific task at hand.

\begin{figure}[th]
    \begin{tcolorbox}[title=User Query, userquery]
    I have a FireFOAM simulation case located at \{case\_path\}.\\
    \{user\_request\}\\
    Always read the contents of a file before modifying it.
    I have compressed the entire case directory, including the README file, into a single long string for you to view and understand my request, as follows: \\
    \{case\_contents\}
    \end{tcolorbox}
    \caption{Case Configuration prompt.}
    \label{fig:case_configuration_prompt}
\end{figure}

\subsection{Job Execution}

Running FireFOAM simulations in a Linux environment can be challenging for entry-level users, particularly since large-scale simulations are often conducted in an HPC environment using a job scheduler like SLURM (\cite{SLURM}), which they are unlikely to have previously used.
Here, we sought to develop a functionality that would allow FoamPilot to handle the execution of simulations. We sought to be able to execute without a scheduler, such as on a local machine or on the head node of an HPC system as a serial job, as this is a common configuration for debugging and testing purposes. We also sought to be able to execute using a job scheduler to run large simulations on numerous multi-core HPC nodes.

When running a FireFOAM simulation locally or on the head node, the agent must prepare the mesh and execute the simulation directly on the command line.
If requested by the user, the agent can also provide a preliminary analysis of the results once the simulation has completed.
When running on an HPC system with a job scheduler like SLURM, the agent must first execute commands like \texttt{scontrol} to identify available resources.
It must then prepare the mesh and determine the size of the mesh. Then, it must perform domain decomposition based on the number of nodes available and the size of the mesh, and finally write a job submission script and submit the job to the queue.

The Job Execution functionality was achieved by providing detailed instructions through prompting, which cause the agent to use the Shell Tool to fulfill the request. The prompts used for serial and HPC jobs are provided in \cref{fig:job_submission_prompt}. For the HPC job, as an additional challenge, the selection of the number of nodes and cores based on the mesh size was left to the agent.

\begin{figure}[ht]
   \centering
    \begin{tcolorbox}[title=User Query, userquery]
    \textbf{Prompt for serial job:} I have a FireFOAM simulation case located at \{case\_path\}. Take a look at the case directory. Mesh the case using the provided script, and then run the simulation in serial on the command line by invoking fireFoam. Write the output to a log file. After the simulation is finished, plot the results of volumetric heat release rate and save them in the case directory. Remember to load environment variables from \{OF\_bashrc\_path\}.\\
    
    \textbf{Prompt for HPC job:} I have a FireFOAM simulation case located at \{case\_path\}. Determine what SLURM queues you have access to. Mesh the case using the provided script. Based on the mesh size and the resources you have available, choose how many nodes to use. Use all physical cores on each node you use. Configure the number of subdomains in the case based on the number of physical cores and decompose the domain. Prepare a SLURM script for the queue and core count, then submit the job. Remember to always load environment variables from \{OF\_bashrc\_path\} before any FOAM command, both in the command line and in the SLURM script. Always read the contents of a file before modifying it. I have compressed the entire case directory, including the README file, into a single long string for you to view and understand my request, as follows: \{case\_contents\}
    \end{tcolorbox}
   \caption{Prompts used to run simulations serially on head node and in parallel using a job scheduler.}
   \label{fig:job_submission_prompt}

\end{figure}

\section{Experimental results and discussion\label{sec:results_and_analysis}}

To ensure consistency across all our experiments, we utilized the same LLM, version, and temperature setting: Azure/OpenAI's GPT-4o, version 2024-05-13, with a temperature setting of $0.0$. The LLM was chosen due to its performance on reasoning tasks and its apparent familiarity with some aspects of the OpenFOAM toolbox.
Despite efforts to maintain reproducibility, the agent still exhibits some variability in success rates when completing tasks with the same user query under identical conditions--a temperature setting of $0.0$ does not guarantee deterministic results. Therefore, each experiment was repeated five times to assess stability. A single user prompt was used in each experiment, and the same prompt was used in each repeat. All of the experiments considered are unambiguous and have a single correct outcome, where a successful outcome is determined by comparing the agent's actions to that of an experienced FireFOAM user. 
The results are summarized in Table \ref{tab:test_results}.

A system prompt was used to define the model's role, behavior, and objectives, and to inform it what tools it has access to, thereby guiding it to produce relevant, accurate, and safe responses. It also helped eliminate redundant content from user queries.
We implemented a system prompt inspired by \cite{structuredChatAgent}, shown in \cref{fig:systemprompt}.

Given the exploratory nature of this project, and given the Shell Tool's ability to execute arbitrary commands, there existed a risk of causing damage to the system on which it was run during our experimentation.
Thus, all experiments were conducted on a dedicated AWS EC2 instance, and HPC jobs were submitted to a dedicated AWS parallelCluster using the SLURM job scheduler. FoamPilot is a small Python-based code, and we employed a cloud-hosted LLM in our experiments, thus the system requirements for the instances were driven entirely by the FireFOAM code.
We note that while our implementation does include a mechanism to allow the user to verify shell commands before their execution, this may still be challenging for users unfamiliar with the Linux command line to judge and manage effectively. 
We would therefore strongly recommend that the testing of agents with shell command execution authority should only be carried out within a secure sandbox environment.

\begin{figure}[h!]
    \centering
    \begin{tcolorbox}[title=System prompt, userquery]
    You are an assistant whose job is to help fire scientists navigate and summarize the source code, modify the simulation case configuration files, and run simulation jobs.
    Respond to the human scientist as helpfully and accurately as possible.
    You have access to the following tools: \{tool\_names\}.
    Use a JSON blob to specify a tool by providing an action key (tool name) and an action\_input key (tool input).
    Valid "action" values: "Final Answer" or \{tool\_names\}.
    Provide only ONE action per \$JSON\_BLOB, as shown:
                
\begin{verbatim}
```
{{
"action": $TOOL_NAME,
"action_input": $INPUT
}}
```
    \end{verbatim}
    \vspace{-0.5cm}
    Follow this format:
   \begin{verbatim}
Question: input question to answer
Thought: consider previous and subsequent steps
while requests is not finished, do
    Action:
    ```
    $JSON_BLOB
    ```
    Observation: action result
end
    \end{verbatim}
    \vspace{-0.5cm}
After the problem is solved, give a final thought to summarize.
    \end{tcolorbox}
    \caption{System prompt, inspired by \cite{structuredChatAgent}.}
    \label{fig:systemprompt}
\end{figure}

Two experiments were considered for the Code Insight functionality, in both of which the agent is prompted to return the mathematical form of an equation that is expressed in the source code. The first case, shown in \cref{fig:code_insight_example}, seeks to obtain the energy equation used in the solid phase, and the second seeks to obtain the form of the non-spherical particle drag force. The agent was consistently successful in both cases, as shown in \cref{tab:test_results}.

The accuracy of the Code Insight functionality depends heavily on whether the RAG Tool retrieves the relevant code files, which is strongly influenced by the precision of the retrieval process. 
The choice of embedding model plays a critical role in determining retrieval accuracy.
We tested several open-source and closed-source embedding models currently available, and some models demonstrated poor performance due to their limited input sizes, resulting in unsatisfactory results. We ultimately chose to produce embeddings using the \texttt{text-embedding-ada-002} model from \cite{openai2022} and created the vector store using the FAISS library (\cite{douze2024faisslibrary}) with LangChain.

We note that while cosine similarity search in embedding space offers significant improvements over manual and automated keyword-based searching, ambiguous user queries can still lead to the retrieval of incorrect code files, causing the agent to produce inaccurate responses.
Popular self-correcting RAG systems are typically based on a multi-agent approach (\cite{yan2024corrective, asai2023self}). 
However, to date no LLM has been trained on the entire source code libraries of OpenFOAM and FireFOAM. 
This raises doubts about whether a secondary LLM in such systems can effectively assess the correctness of the retrieved information.
This limitation primarily arises from the current dependence of LLMs on labeled data for training. 
Acquiring such labeled data, such as well-commented source code for a large-scale open-source simulation software package, is prohibitively expensive.
Still, agent performance could likely be significantly enhanced if it were further trained on the OpenFOAM and FireFOAM source code using label-free methods, such as continued pretraining.

\begin{figure}[h!]
    \centering
    \begin{tcolorbox}[title=User Query, userquery]
        Find the source code related to the 1D reacting solid model. Using the model's definition of the energy equation, write the energy equation in Latex.
    \end{tcolorbox}
    
    \begin{tcolorbox}[title= Agent response, agentcall] 
    **Possible File Locations:** 
    
    [fireFoam/lib/regionModels/pyrolysisModels/reactingOneDimFF/reactingOneDimFF.C, 
    fireFoam/lib/regionModels/pyrolysisModels/complexFuelPyrolysis/complexFuelPyrolysis.C]\\
    
    \textit{(Retrieval and analysis of the source code)}\\
    
    **Summary:** 
    
    The energy equation in the \texttt{reactingOneDimFF} class is given by:
    \begin{align*}
    \frac{\partial (\rho h)}{\partial t} &+ \nabla \cdot (\phi_{\text{pyrolysis}} h) 
    - \nabla \cdot (\alpha \nabla h) 
    + \nabla \cdot (\alpha \nabla h) 
    - \nabla \cdot (\kappa \nabla T) \\
    &= \text{chemistryQdot} + \text{solidChemistry->RRsHs()}
    \end{align*}
    
    If \texttt{gasHSource} is true, an additional term is added to the equation:
    
    \[
    + \nabla \cdot (\phi_{\text{Gas}})
    \]
    
    And if \texttt{qrHSource} is true, another term is added:
    
    \[
    + \nabla \cdot (\phi_{\text{qr}} \cdot n\text{MagSf})
    \]
    
    This equation is solved in the \texttt{solveEnergy} function within the \texttt{reactingOneDimFF} class, located in the \texttt{reactingOneDimFF.C} source file.
    \end{tcolorbox}
    \caption{Code Insight example: Summarizing the 1D solid-phase energy equation based on FireFOAM source code. Note that the appearance of apparently canceling diffusion terms on the left-hand-side is correct; the terms appear this way in the code due to details of the numerical implementation.} 
    \label{fig:code_insight_example}
\end{figure}

Considering the Case Configuration functionality, two examples were tested based on the tutorial cases provided with FireFOAM: modification of the size of a burner, and modification of the thickness of cardboard boxes. The result of the former case is shown in \cref{fig:case_configuration}, where we see that the agent correctly identifies both files requiring modification and modifies them accurately. In the latter case, the agent only needed to modify one file, but the task was more difficult since `cardboard thickness' is not a keyword present anywhere in the case files.
In both cases, we see in \cref{tab:test_results} that the agent was consistently successful.
We note that only a limited number of parameters need to be modified between the two examples, making them relatively simple cases. In our experiments, we observed that case modifications significantly more complex than those shown failed consistently. 
We also note that intermediate and advanced users may be able to detect errors in the configured cases that do not cause FireFOAM to crash, but rather cause unintended outputs; however, less experienced users would struggle to do so. 

\begin{figure}[h!]
    \begin{minted}[bgcolor=bg, frame=lines, fontsize=\small, escapeinside=||]{python}
    # system/topoSetDict
    # Original
    box (-0.15 -0.15 -0.001) (0.15 0.15 0.001);
    # Modified 
    box (-0.3 -0.3 -0.001) (0.3 0.3 0.001);
    
    # system/snappyHexMeshDict
    # Original
    min (-0.15 -0.15 0.0 );
    max ( 0.15  0.15 0.0 );
    # Modified
    min (-0.3  -0.3  0.0 );
    max ( 0.3   0.3  0.0 );
    \end{minted}
    \caption{Case Configuration example: Modifying burner size in FireFOAM's \texttt{poolFireMcCaffrey} tutorial case from 0.3m to 0.6m.}
    \label{fig:case_configuration}
\end{figure}

In the case of the Job Execution functionality, we ran two series of tests: running a serial job on the head node, and submitting a job to a scheduler. We observed in general that the Job Execution functionality was mostly unsuccessful unless given a highly detailed prompt as in \cref{fig:job_submission_prompt}. In the case of the serial job, the agent was successful in three of five runs, requiring seven agent-tool loops to complete the task as prescribed in \cref{fig:job_submission_prompt}. In one case, the agent incorrectly called the RAG Tool instead of the Shell Tool, which polluted the context window with a significant amount of irrelevant information. Still, the agent recovered and successfully ran the serial job with the Shell Tool and plotted the results using the Python Interpreter Tool in 17 loops. In the failing run, the agent hallucinated the existence of a number of files in the case directory, and ultimately failed to recover.

The agent performed much more poorly in the HPC tests, only succeeding once in nine agent-tool loops. In one failing case, the agent failed to correctly identify the HPC environment and issued incorrect SLURM commands. In the other failing cases, the agent performed almost all sub-tasks correctly per the system prompt, but failed to correctly setup the environment in its job submission script. We observed that the agent consistently gave reasonable estimations for the number of cores to use for the simulation based on the mesh it generated.

A more challenging task involves handling multi-functionality queries. 
These tasks typically require the agent to iteratively call multiple tools based on the user's query. 
An example of such a task would be modifying a simulation case file based on information retrieved from the source code.
Compared to simply perturbing parameter values, the agent is expected to review the C++ code retrieved by the RAG Tool and, based on this, identify the correct case files in which to replace specific keywords or even functions.
We report one such test we conducted, where we prompted the agent to change the drag force model on the droplets in FireFOAM's \texttt{burningBoxSuppression} tutorial case without specifying how such a modification should be achieved, thereby combining Code Insight and Case Configuration functionalities.
The agent was successful in two runs, requiring only three agent-tool loops to use both the RAG Tool and the Shell Tool to learn what modifications needed to be made and make the necessary modifications correctly. In one run, the agent's RAG Tool query was incorrect and returned irrelevant information, from which it did not recover. In the other two runs, the agent successfully used the RAG Tool to retrieve relevant files, but failed to glean the necessary information from them to make the requested modification to the simulation configuration, and instead made incorrect modifications.
We note that on multi-functionality tests that were more complex than the relatively simple one presented here, the agent failed consistently.

\renewcommand{\arraystretch}{1.1}
\begin{table}[h!]
\centering
\begin{tabular}{lll}
\textbf{Functionality} & \textbf{Task} & \textbf{Success rate} \\ \hline
Code Insight      & Solid energy equation     & 5/5      \\
Code Insight      & Non-spherical drag equation    & 5/5      \\
Case Configuration      & Burner size modification      & 5/5     \\
Case Configuration      &  Cardboard thickness modification     &  5/5      \\
Job Execution      & Run serial job      & 4/5      \\
Job Execution      & Submit HPC job      & 1/5      \\
Code Insight \& Case Configuration      & Modify droplet drag model      & 2/5      \\
\end{tabular}
\vspace{0.2cm}
\caption{Summary of test results.}
\label{tab:test_results}
\end{table}

\section{Conclusions and future work}

An LLM agent, FoamPilot, was developed as a proof-of-concept with the aim of reducing the complexity and time required for source code navigation, simulation setup and simulation execution, thereby making fire dynamics simulations more accessible and efficient for both new and experienced users. In our exploration, we found that the agent was consistently successful for tasks of low complexity, but that its success rate dropped precipitously with increasing task complexity. 

Important functionalities that were not addressed in this work include the ability to run simulations asynchronously, which requires the LLM agent to save and recover its state between sessions. This would be particularly useful to users that are less familiar with the HPC environment and workflow. In addition, a robust ability for optional human feedback during FoamPilot's operations was not achieved due to technical challenges with the chosen agentization framework. A request for human approval of any tool usage was included as a necessary safety precaution during testing; however, challenges arose in the LangChain/LangGraph framework with respect to the solicitation and inclusion of substantial human feedback between tool usages by the LLM agent. 

Our experiments considered only GPT-4o for the agent's LLM, and it is plausible that there are LLMs available at the time of writing that would have achieved better results in our experiments. We anticipate greater success for case configuration tasks when reliably-structured LLM outputs are combined synergistically with non-AI case configuration tools that embed expert knowledge. We expect that near-future LLMs specialized for reasoning tasks will perform better on the multi-step, complex tasks required for setting up and executing simulations. Furthermore, we expect that present and future LLM-enhanced developer tools, such as GitHub CoPilot and Cursor AI Code Editor, will certainly outperform our implementation of the Code Insight functionality, since this functionality is useful for all software developers, not just FireFOAM developers. Indeed, the challenges identified and the solution approaches developed in this work are not unique to FireFOAM, they are applicable to many scientific simulation workflows. The development of generalized frameworks for LLM agents to interact with and control simulation software will be beneficial.

Looking ahead, we note that an increasing number of large language models are becoming multimodal, capable of processing image data as input. We believe this capability could improve the agent's understanding of simulation configurations if the geometry and mesh are presented visually to provide further context regarding the simulation at hand.

Lastly, we found that the limited domain-specific knowledge of general-purpose LLMs reduces FoamPilot's ability to handle complex tasks using FireFOAM, particularly those involving multiple functionalities. 
We note that FireFOAM's codebase is itself around 1M tokens, whereas the OpenFOAM toolbox on which it is built is approximately 10x larger. Thus, FireFOAM, or large parts of it, may fit in the expanded context windows of future LLMs. Additionally, continued pre-training on the OpenFOAM or combined FireFOAM/OpenFOAM codebase may potentially allow accurate zero-shot prompting for analyzing and supporting further code developments.

\clearpage

\bibliographystyle{unsrtnat}  
\bibliography{references}

\end{document}